# Emergent Linguistic Rules from Inducing Decision Trees: Disambiguating Discourse Clue Words
## (To Appear in AAAI Proceedings, 1994)

### Eric V. Siegel and Kathleen R. McKeown
Department of Computer Science
Columbia University
New York, NY 10027
evs@cs.columbia.edu, kathy@cs.columbia.edu

## Abstract

We apply decision tree induction to the problem of discourse clue word sense disambiguation. The automatic partitioning of the training set which is intrinsic to decision tree induction gives rise to linguistically viable rules.

## Introduction

*Discourse clue words* function to convey information about the topical flow of a discourse. Clue words can be used both to bracket discourse segments and to describe the discourse relationship between these segments. For example, *say* can introduce a set of examples, as in "Should terrestrial mammals be taken in the same breath as types of mammals as *say* persons and apes?"[1] However, each word in Table 1 has at least one alternative meaning where the word contributes not to discourse level semantics, but to the semantic content of individual sentences; this is termed its **sentential** meaning. For example, *say* can mean "To express in words", as in "I don't want to *say* that he chickened out at a presentation or anything but he is in Toronto..." Therefore, to take advantage of the information supplied by discourse clue words, a system must first be able to disambiguate between such a word's **sentential** and **discourse** senses.[2]

In this paper, we perform automatic decision tree induction for this problem of discourse clue word disambiguation using a genetic algorithm. We show several advantages to our approach. First, the different decision trees that result encode a variety of linguistic generalizations about clue words. We show how such linguistic rules emerge automatically from the *training set partitioning* which occurs during decision tree induction. These rules can be examined in order to evaluate the validity of induced decision trees. Examining the rules also provides insights as to the type of syntactic information necessary to further improve clue word sense disambiguation. Second, decision trees are induced which generalize across a set of 34 clue words (see Table 1) in contrast to previous automated approaches to word sense disambiguation which typically have focused on discriminating the senses of one word at a time [Schuetze 1992] [Brown et al 1991] [Leacock et al 1993] [Black 1988] [Grishman and Sterling 1993] [Yarowsky 1993]. As we show, this allows for greater learning potential than dealing with words individually. There are some problems with the domain of disambiguation for clue words and we discuss these, indicating why our approach is likely to be more helpful for other disambiguation problems.

The following four sections discuss previous work on disambiguation, describe our approach, present experimental results in both linguistic and numerical terms, and draw conclusions and present our future research directions.

## Previous Work

Hirschberg and Litman [1993] explore several methods for disambiguating clue words, including measuring the ability with which this task can be performed by looking only at the punctuation marks immediately before and after a clue word, suggesting the strategy embodied by the decision tree in Figure 1.[3] This small decision tree classifies clue words as **discourse** exactly when there is a period or a comma immediately preceding, and as **sentential** in all other cases. This means, for example, that a word is classified as **discourse** when it is the first word of a sentence. For such a simple strategy, the decision tree performs to a relatively high degree of accuracy over our corpus: 79.16%. Our work investigates disambiguation strategies for clue words which involve looking at near-by words in addition to punctuation marks.

The automatic acquisition of disambiguation strategies has been applied to many types of ambiguity problems, including word sense disambiguation [Schuetze 1992] [Leacock et al 1993] [Yarowsky 1993] [Brown et al 1991], determiner prediction [Knight forthcoming], and several parsing problems [Resnik 1993] [Magerman 1993]. Previous work using decision tree induction for disambiguation includes work by Black [1988] (word sense disambiguation), Knight [forthcoming] (determiner prediction), Resnik [1993] (coordination parsing) and Magerman [1993] (syntactic parsing). Automatic approaches to word sense disambiguation have thus far primarily focussed on disambiguating one word at a time.

---

[1] The examples come from the corpus used in this study.

[2] See Hirschberg and Litman [1993] and Schiffrin [1987] for details on other clue words and more information about clue words in general. In this paper, *clue word* refers to a word from Table 1, regardless of the particular sense with which it occurs.

[3] This decision tree is a slightly simplified extrapolation of Table 11 from Hirschberg and Litman [1993]. Hirschberg and Litman [1993] also investigated the ocurrence of clue words adjacent to one another, but with no conclusive results.

Table 1: Discourse clue words and the fraction of times each is used in its **discourse** sense.

| Clue word | Fraction | Clue word | Fraction | Clue word | Fraction | Clue word | Fraction |
|---|---|---|---|---|---|---|---|
| *and* | 137/348 | *see* | 0/29 | *no* | 0/9 | *next* | 0/4 |
| *now* | 64/75 | *actually* | 1/29 | *although* | 5/9 | *yes* | 0/3 |
| *so* | 55/74 | *first* | 0/25 | *indeed* | 5/8 | *since* | 0/3 |
| *like* | 6/71 | *also* | 5/20 | *OK* | 8/8 | *except* | 0/3 |
| *but* | 27/56 | *then* | 6/15 | *however* | 8/8 | *therefore* | 0/2 |
| *or* | 17/55 | *further* | 8/14 | *generally* | 1/6 | *otherwise* | 1/1 |
| *say* | 25/36 | *finally* | 11/11 | *similarly* | 3/5 | *anyway* | 0/1 |
| *well* | 13/35 | *right* | 0/10 | *basically* | 1/5 | | |
| *look* | 0/35 | *because* | 0/10 | *second* | 0/4 | | |

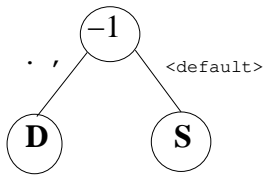

Figure 1: Manually created decision tree with accuracy 79.16%.

## Approach

In this study, we expand on the orthographic approach to clue word disambiguation described by Hirschberg and Litman [1993] by allowing decision trees to test not only for adjacent punctuation marks and clue words, but also for near-by words of any kind, and by allowing the decision trees to discriminate between clue words. The set of *attributes* available to a decision tree are the *tokens* (words and punctuation marks) appearing immediately to the left of the ambiguous word, immediately to the right of the ambiguous word, and 2, 3, and 4 spaces to the right of the ambiguous word, as well as the ambiguous word itself (*attribute* **0**), that is, {**-1, 0, 1, 2, 3, 4**}. This set of *attributes* were selected to test whether the decision trees would find a wider window of *tokens* useful for clue word disambiguation, but, as was automatically determined, only the adjacent *tokens* and the ambiguous word itself were deemed useful. No information describing syntactic structure is explicitly available to decision trees. The genetic algorithm determines automatically which words or punctuation in these positions are important for disambiguation.

### Decision Trees

Figures 2 and 3 show example decision trees which were automatically induced for clue word sense disambiguation. Internal nodes are labeled with *token positions*

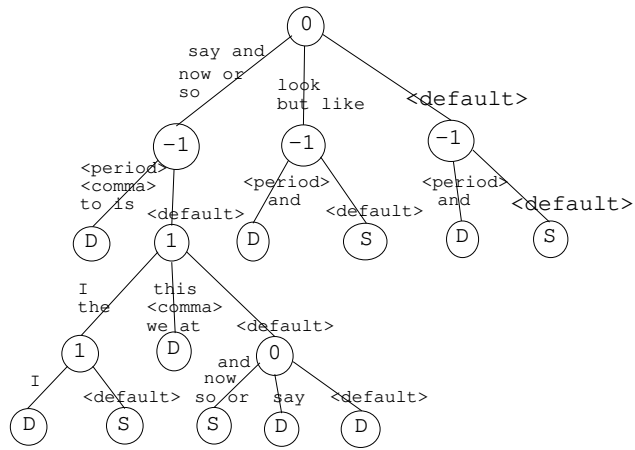

Figure 2: Decision tree automatically induced by the genetic algorithm. This tree disambiguated with 81.10% accuracy over the training set, and with 82.30% accuracy over the test set.

(*attributes*), arcs are labeled with sets of *tokens* (*values*), and leaves are labeled with *classes*, that is, either **discourse** or **sentential**. Given a text fragment containing a clue word, a decision tree classifies the word as to its sense by a deterministic traversal of the tree, starting at the root, down to a leaf. During traversal, an arc descending from the current (internal) node is selected in order to continue the traversal. This arc is chosen by finding the first descending arc, going from left to right, containing the token at the text fragment position indicated by the current node's label. For example, to traverse the tree in Figure 2, starting at the root node, the leftmost arc is traversed if the word at position **0** is one of the words on the arc (e.g., *say*). The rightmost arc under each internal node is labeled "default", and is traversed when none of its sister arcs contain the correct token.

In order to increase the likelihood that an induced decision tree will embody valid generalizations, as opposed to being over-fitted to the particular set of training examples, only the tokens which appear with frequency above a threshold of 15 in the training cases are permitted in the value sets of a decision tree (see the subsection "The Training Data" for details on the training corpus),

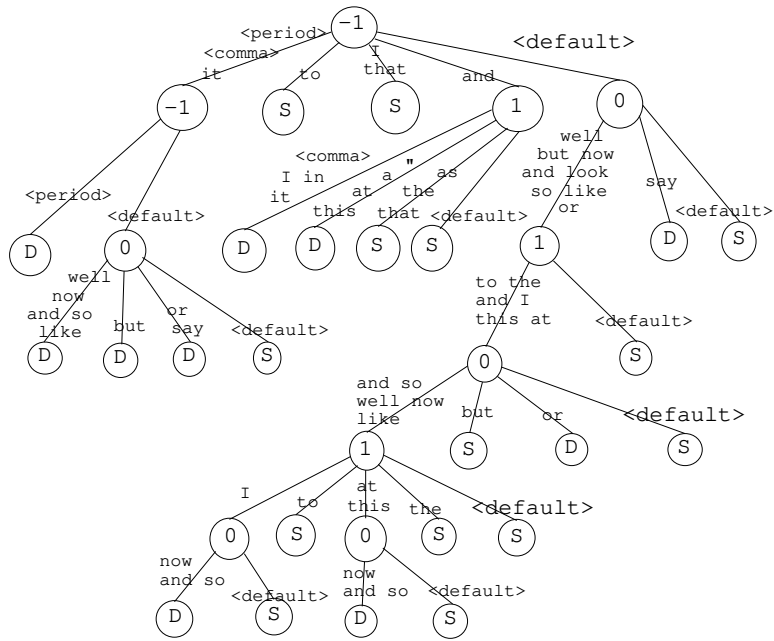

Figure 3: Decision tree automatically induced by the genetic algorithm. This tree disambiguated with 84.99% accuracy over the training set, and with 82.30% accuracy over the test set.

specifically:[4]

{<period>, <comma>, <apostrophe-s>, *a*, *and*, *are*, *as*, *at*, *can*, *for*, *I*, *in*, *is*, *it*, *of*, *that*, *the*, *this*, *to*, *we*, *you*}

A separate set of tokens is available to the arcs under nodes labeled **0**, namely the discourse clue words which appear with frequency greater than 4 in the training cases. (Only clue words appear at position **0**.) This threshold was chosen to allow infrequent clue words to be specified by a decision tree, but to still avert overfitting to the training data.

## Decision Tree Induction

The corpus used in this study supplies 1,027 examples. Table 2 shows sample data. Each training case has a manually specified class, and a value corresponding to each of 6 attributes. In order to predict the performance of an induced decision tree over unseen data, the induction procedure is run over a random half of the corpus (the *training set*), and the resulting decision tree is then evaluated over the remaining half of the corpus (the *test set*). This division of the data is performed randomly before each run.[5]

The induction procedure used in this study is a genetic algorithm (GA) [Holland 1975], a weak learning method which has been applied to a wide range of tasks in optimization, machine learning, and automatic computer program induction [Goldberg 1989] [Koza 1992]. Inspired by Darwinian survival of the fittest, the GA works with a pool (*population*) of *individuals*, stochastically performing *reproductive operators* on the individuals, depending on some notion of *fitness*. Reproductive operators include *crossover*, a stochastic procedure by which two individuals are combined to create a third, and *mutation*, by which an individual undergoes a random alteration. In our work, individuals are both decision trees and the token sets which correspond to decision tree arcs. Fitness corresponds to the number of training cases correctly classified by a decision tree. The GA outputs the highest fit decision tree it encounters. Siegel [1994] describes the details of GA decision tree induction applied in this work. Subsection "Numerical Results" in this paper contrasts GA decision tree induction to classical decision tree induction techniques.

## The Training Data

The 1,027 training examples come from a corpus used by Hirschberg and Litman [1993]. This is a transcript of a single speaker speech, preceeded by introductory remarks by other speakers, in which each occurrence of the words in Table 1 has been manually marked as to its meaning by a linguist.[6] When marking the corpus, the linguist had access to the entire transcript plus a recording of the speech. Therefore, much more information was available to the linguist than there is to a decision tree. Regardless, about 7% were deemed ambiguous by the linguist. The "ambiguous" examples were left out of this study since they provide no information on how to disambiguate. 407 of the 1,027 unambiguous lexical items (39.63%) were marked as **discourse**, and 620 (60.37%) were marked as **sentential**. See Hirschberg and Litman [1993] for more detail on the corpus and the distribution of data within it.

Table 2: Example training cases.

| -1 | 0 | 1 | 2 | 3 | 4 | Class |
|---|---|---|---|---|---|---|
| . | But | we | stop | there | because | D |
| . | Now | that | doesn't | mean | we | D |
| to | look | more | like | sentences | . | S |
| , | and | that's | on | the | second | S |

---

[4] Tokens are case-insensitive (capitalization doesn't matter), but inflection-sensitive (*a* is different than *an*).

[5] Because of this random division, the frequency distribution of tokens in the training set varies, so the valid token and clue word sets actually varies slightly.

# Results

Since the division between training and test cases is random for each run, and since the GA is a stochastic method, each run of the GA gives rise to a unique decision tree.[7] We performed 58 runs, thus generating 58 trees. We evaluate these trees in two ways. First, by manually examining several high scoring trees, we show they yield linguistically valid rules. Second, we measure the average performance of induced decision trees.

## Linguistic Results

The small decision tree in Figure 1, a tree obtained manually by Hirschberg and Litman [1993], yields an accuracy of 79.16%. To attain any improvement in accuracy, a more complex *partitioning* of the training cases must take place, by which the GA focuses on the cases where the majority of error lies. It is by this partitioning process that additional linguistic rules are induced.

A decision tree implicitly partitions the training (and test) cases; each rule embedded in a decision tree corresponds to a partition. As an example, the small decision tree in Figure 1 corresponds to the following simple partitioning of the training data:

-1 = <period> is true for 189 cases (185 **discourse**).
-1 = <comma> is true for 72 cases (42 **discourse**).
766 cases remain (180 **discourse**).

In order to attain a higher accuracy than that of the small decision tree, the partition consisting of the 766 "remaining" cases, for example, is a viable candidate for re-partitioning – rules must be found which apply to subpartitions of that partition. As we show here, many of the induced rules tend to be linguistically viable.

There are two ways to examine the resulting rules. First, we identify general rules that apply to sets of clue words (i.e., more than one) from several trees. In particular, we note that different trees yield different generalizations. Second, we identify all generalizations encoded in high scoring trees for individual clue words. These generalizations identify the rules that, in combination, can be used for a single clue word. In analyzing these generalizations, we note where they are specific to the corpus and where we expect them to generalize to different domains.

**Multiple Clue Word Rules.** Table 3 displays example linguistic rules extracted from various decision trees, and lists the clue words to which they apply. Each rule consists of a comparison (under column "If") and the clue word sense which results if the comparison holds (under column "Then" – "S" stands for **sentential** and "D" stands for **discourse**). The "Linguistic Template" column indicates the most frequent part of speech of the clue word when the comparison holds, as determined manually, and is elaborated below. "Accuracy" shows the number of cases in the corpus for which the rule holds, divided by the number of cases in the corpus which match the pattern.

These rules strongly suggest strategies by which part of speech is used for disambiguation; the rules embody the fact that a clue word's sense is **sentential** if its part of speech is not a conjunction, and must be further disambiguated if it is a conjunction.

The first rule classifies an occurrence of either *see, look, further* or *say* as **Sentential** if position -1 is *to*. (These are the clue words for which this rule holds in the corpus.) Of the 30 times for which this condition holds, the rule is correct 29; the rule holds exactly when the listed words are behaving as verbs, as indicated by the linguistic template "*to* <verb>", e.g.:

*...we can foster this integration of AI techniques and database technology to* **further** *the goal of integrating the two fields into Expert Database Systems.*

This example is in fact the only occurrence of *further* in the corpus for which *to* is the immediately preceding token. However, the GA can induce this rule since it is generally applicable over the 4 clue words (as shown in the tree of Figure 3). This demonstrates the benefit gained by simultaneously disambiguating multiple words.

The second rule listed (100% accuracy) embodies two different "syntactic templates". Both are detected by checking for **-1 =** *the*. The first, which occurs for *like, and* and *right*, determines that the sense is **sentential** if the clue word is being used as a noun[8], as in:

*...a lot of work going on now in what's called non-monotonic reasoning, circumscription and the* **like**...

and the second, which occurs for *right, first* and *next*, determines that the sense is **sentential** if the clue word is being used as an adjective in a noun phrase, as in:

---

[6]We used one linguist's markings, whereas Hirschberg and Litman [1993] used and correlated the judgements of that and another linguist, discarding those cases in which there was disagreement. Thus, the data we used was slightly more noisy than that used by Hirschberg and Litman [1993]. Further, we used a slightly larger portion of the marked transcript than is reported on by Hirschberg and Litman [1993].

[7]Technically, there is a very small possibility that the same decision tree will be induced by two different runs of the GA.

[8]*And* is a noun when it is used to refer to the logical operator.

*...I think this is the **first** time those three are cooperating...*

The third rule (90.11% accuracy) pinpoints the collocation *"as well"*. When in this collocation, *well* is being used as an adverb, e.g.:

*We could have just as **well** done without it but the system would run a lot more slowly.*

The fourth rule (76.92% accuracy), which applies to the 8 clue words listed, approximates the cases where a clue word is being used as an adverb, as in:

*And then in the summer of 1985 Ron left the West Coast to travel east to New Jersey where he is **now** at AT&T Bell Laboratories as head of the AI Principles Research Department.*

However, the condition "-1 = *is*" holds for some cases in which a clue word is used in its **discourse** sense, as in:

*...and the second question is **well** where do we stop.*

Therefore, this particular rule is too simplistic for some cases. However, it has indicated for us a disambiguation method which uses the part of speech of the clue word.

**Single Clue Word Rules.** Table 4 shows the way the decision trees in Figures 2 and 3 disambiguate *and* and *say*, respectively. The decision trees are explicitly broken down into the rules used to disambiguate the individual clue words. The columns in the table are the same as the previous table, with the addition of "Decision tree", which points to the tree being analyzed. The rules for each word are listed in the order in which they are considered when traversing the decision tree. Therefore, for example, the condition of the fourth rule for *and* is only tried on cases for which -1 is none of <period>, <comma>, or *is*, and this is reflected in the number of cases for which the condition holds, as listed in the "Accuracy" column. This number of occurrences is a count across the entire corpus; that is, both the training and test cases. The overall accuracy with which the example decision trees disambiguate the individual clue words is also shown.

The rules for *and* reflect the fact that, when coordinating noun phrases, *and* is usually being used in its **sentential** sense, and, when coordinating clauses, *and* is most often being used in its **discourse** sense.

The first two rules for *and* are the same as the first two rules of the small decision tree in Figure 1. The third and eighth rules hold for too few examples to draw any conclusions. The condition of the fourth rule approximates the cases for which *and* is being used to coordinate noun phrases, since most definite noun phrases are not the subject of a clause in the corpus. For example:

*...I've been very lucky to be able to work with Don Marshand **and** the institute in organizing this...*

This is clearly too simple a strategy (64.29% accuracy), but provides insight for improved strategies.

The fifth, sixth and seventh rules (75.00%, 85.71% and 83.33% accuracy) approximate the cases for which

*and* is coordinating clauses, since *I*, *we* and *this* are most frequently the subject of a clause in the corpus, as in:

*The idea of the tutorial sessions was precisely to try to bring people up to speed in areas that they might not be familiar with **and** I hope the tutorials accomplish that for you.*

The small tree of Figure 1, which disambiguates in general with accuracy 79.16%, only disambiguates *and* with accuracy 71.84% (The small tree disambiguates the occurrences of clue words other than *and* with accuracy 82.92%). However, the overall accuracy with which the decision tree in Figure 2 disambiguates *and* is 76.44%.

The decision tree in Figure 3 treats *say* differently and separately from the other clue words: After the first default arc is traversed, *say* is always disambiguated as **discourse**, while other words are treated differently (e.g., *well* is further tested). This demonstrates the utility of allowing decision trees to discriminate between clue words, since *say* occurs with sense **discourse** more frequently than most other clue words; *say* is only disambiguated with accuracy 41.67% by the small tree of Figure 1, but is disambiguated with accuracy 83.33% by the induced decision tree of Figure 3.

The cases for which the tree in Figure 3 classifies *say* as **sentential** are when *say* behaves as a verb, as in:

*That is if I **say** that John is both a Quaker and a Republican...*

As demonstrated by the contents of Table 4, most instances of clue words in the corpus are disambiguated by rules which hold with high accuracy, as measured across the entire corpus, while the decision trees were induced over only half of the corpus (the training set). This indicates that performance will remain high for unseen examples from similar corpora.

## Numerical Results

From 58 runs of the GA, each with a random division between training and test cases, the maximum score over the test cases was 83.85%.[9] The performance of such a tree over unseen data ideally requires further formal evaluation with more test data.

The average score over the test cases for the 58 runs was 79.20%. The average disparity between training and test scores, 2.64, is not large. Therefore, the rules of induced decision trees tend to perform well over unseen data, although it is inconclusive whether their combined contribution to disambiguation accuracy improves over the overall performance of the small tree in Figure 1 (79.16%) for the entire set of clue words. However, decision trees clearly aid in the disambiguation of several of the clue words, e.g. *say* and *and*.

These results reflect the difficulty inherent to the task of clue word sense disambiguation. Hirschberg and Lit-man [1993] report that 7.87% of the examples manually marked by the authors were either disagreed upon by the authors, or were decidedly ambiguous.

---

[9]This is the maximum test score of the decision trees which performed the best of their run over the *training* cases. This same pool of trees is considered for average test performance.

Table 3: Linguistic rules extracted from various automatically induced decision trees.

| Clue words | Rule If | Rule Then | Linguistic template | Accuracy |
|---|---|---|---|---|
| see, look, further, say | -1 = to | S | to <verb> | 29/30 = 96.67% |
| like, and, right right, first, next | -1 = the | S | the <noun> the <adj> | 18/18 = 100.00% |
| well | -1 = as | S | as well | 10/11 = 90.11% |
| also, now, generally actually, basically | -1 = is | S | is <adverb> | 10/13 = 76.92% |

Table 4: The rules used by sample trees to disambiguate *and* and *say*.

| Clue word | Decision tree | Rule If | Then | Linguistic template | Accuracy |
|---|---|---|---|---|---|
| and | Figure 2 | -1 = <period> | D | sentence initial | 29/30 = 96.67% |
| | | -1 = <comma> | D | clause initial | 18/25 = 72.00% |
| | | -1 = is | D | (inconclusive) | 1/1 = 100.00% |
| | | 1 = the | S | and <def NP> | 9/14 = 64.29% |
| | | 1 = I | D | and <subject> | 9/12 = 75.00% |
| | | 1 = we | D | and <subject> | 6/7 = 85.71% |
| | | 1 = this | D | and <subject> | 5/6 = 83.33% |
| | | 1 = at | D | (inconclusive) | 1/2 = 50.00% |
| | | else (default) | S | | 188/251 = 74.90% |
| | | | | Overall accuracy for *and*: | 266/348 = 76.44% |
| say | Figure 3 | -1 = to | S | to <verb> | 4/4 = 100.00% |
| | | -1 = I | S | I <verb> | 2/2 = 100.00% |
| | | else (default) | D | | 24/30 = 80.00% |
| | | | | Overall accuracy for *say*: | 30/36 = 83.33% |

Many disambiguation tasks will presumably not have a simple strategy (such as that embodied by the small decision tree in Figure 1) which performs to such a high degree of accuracy. For example, the aspectual classification of a clause requires the interaction of several syntactic constituents of the clause [Pustejovsky 1991]. Therefore, since the disparity between training and test performance is moderate, decision tree induction is likely, in general, to outperform such simple strategies for disambiguation tasks.

As a benchmark, several top-down (*recursive partitioning*) decision tree induction methods [Quinlan 1986] [Breiman et al 1984] were applied to the disambiguation corpus.[10] This comparison was motivated by the fact that top-down decision tree induction is the more established method for decision tree induction. The best top-down method disambiguated the test cases with accuracy 79.06% on average (based on 200 runs, each with a random division between training and test sets), which is comparable to the GA's average performance, 79.20%.

GAs are a weak learning method, which often require less explicit engineering of heuristics than top-down induction. For an investigation of the generalization performance of GA decision tree induction see Siegel [1994].

[10] These experiments were performed using the IND decision tree induction package [Buntine and Caruana 1991].

Tackett [1993] and Greene & Smith [1987] have also performed comparisons between GA techniques and recursive partitioning methods.

# Conclusions and Future Work

The disambiguation of *and* and *say*, as well as other clue words, has benefited from the integration of knowledge about surrounding words, without the explicit encoding of syntactic data. Further, we have demonstrated that the automatic partitioning of the training set during decision tree induction provides an array of linguistically viable rules. These rules provide insights as to syntactic information which would be additionally beneficial for clue word sense disambiguation. Further, the rules can help linguists evaluate the validity of induced decision trees.

We have demonstrated the utility of disambiguating a set of words simultaneously: generalizations which apply over several words are induced, and, when training over a small corpus, this allows generalizations to be made on examples that occur extremely infrequently (e.g., once).

We plan to apply machine learning methods to aspectual ambiguity. The aspectual class of a clause depends on a complex interaction between the verb, its particles, and its arguments [Pustejovsky 1991]. Induction will be performed simultaneously over a set of verbs, with access to the syntactic parse of example clauses.

## Acknowledgments

We wish to thank Diane Litman and Julia Hirschberg for generously providing the marked transcript of spoken English used in this work. We also thank Diane Litman for extremely valuable feedback regarding this work. Additionally, thank you Rebecca J. Passonneau, Jacques Robin and Vasileios Hatzivassiloglou for many important comments and suggestions. This work was partially supported by ONR/ARPA grant N00014-89-J-1782 and NSF grant GER-90-2406.